\documentclass{article}

 \usepackage[dblblindworkshop, final]{neurips_2026}
\workshoptitle{AI for Stochastic Dynamics}

\usepackage[utf8]{inputenc} 
\usepackage[T1]{fontenc}    
\usepackage{hyperref}       
\hypersetup{hidelinks}
\usepackage{url}            
\usepackage{booktabs}       
\usepackage{amsfonts}       
\usepackage{nicefrac}       
\usepackage{microtype}      
\usepackage{xcolor}         

\usepackage{graphicx}
\usepackage{amsmath}

\newcommand{\RMS}{\operatorname{RMS}}

\title{Cross-Time Directional Selection in Diffusion Sampling}

\author{%
  Dhia Naouali
  \\ Independent Researcher  \\
  \texttt{dhia12naouali@gmail.com} \\
}

\begin{document}

\maketitle

\begin{abstract}

  How strongly do the remaining diffusion-sampling steps amplify a perturbation at a late latent state? Standard measurements answer this question with newly sampled isotropic noise, even though perturbations encountered during sampling have already been transformed by earlier steps. We compare these two cases directly. For each trajectory, we transport a centered  perturbation from an earlier step to a late state, then replay its direction at the same magnitude as a newly sampled isotropic perturbation; both then undergo the same remaining
  updates. Across the samplers we study, median paired shaped-to-fresh angular-gain ratios range from $1.14$ to $2.35$. Shaped angular gain exceeds its matched
fresh counterpart in every trajectory in the original main cohorts. The same pattern appears when endpoint change is measured by latent RMS. The effect also appears in perceptual feature representations: earlier-sampling directions cause larger feature
  changes at the endpoint, even when the corresponding pixel-space change is comparable. Permuting shaped directions across trajectories weakens the effect, including within class, indicating that the advantage depends on alignment with the receiving trajectory as well as on shared directional structure.

\end{abstract}

\section{Introduction}
\label{sec:intro}

Perturbation tests estimate how an error during diffusion sampling changes the final sample. The result depends on when the error enters the trajectory, its magnitude, and its direction. A newly sampled Gaussian
perturbation is isotropic, whereas an error carried forward from an earlier step has been transformed by the intervening updates. Whether these two kinds of perturbation have the same effect on the final sample is an empirical question.

Prior work has identified distinct temporal regimes in diffusion, regularity in sampling paths, the
accumulation of errors, and anisotropic score geometry \citep{biroli2024,chen2024,livanderschaar2024,ventura2025}. Together, these findings raise a more specific question: do early sampling
steps produce perturbation directions that the remaining sampler amplifies particularly strongly? Averaging over newly sampled probes cannot resolve this question, because it does not test directions generated by the trajectory itself. Likewise, the alignment of a trajectory with its endpoint describes the path taken by one sample, rather than the sampler's response to nearby states.

We compare newly sampled perturbations with directions produced upstream, while holding the remaining updates fixed. Starting from a Gaussian direction, we perturb an earlier state in opposite directions and propagate both perturbed trajectories to a common late boundary. Their centered difference defines the \emph{upstream-shaped} direction, which we normalize and replay around the unperturbed boundary state.
The fresh probe uses the original Gaussian direction at the same RMS magnitude. This construction holds the perturbation size and late reference state fixed, leaving the direction produced by upstream transport as the only difference between the two probes.


Across the three samplers, upstream-shaped directions have median paired
angular gains $1.14$--$2.35$ times those of matched fresh probes. Shaped
angular gain exceeds fresh gain in every trajectory in the original main
cohorts. The ordering persists over a shorter late window and when measured
by latent-RMS gain. Repeating the SiT comparison over an eightfold range
of small replay magnitudes leaves the median paired ratio essentially
unchanged. At equal replay magnitude around the endpoint state, early-shaped directions also induce larger changes in two independent image-feature representations.

We then test whether the effect transfers across trajectories by replaying shaped directions on a different trajectory. In new SiT and DiT cohorts, a trajectory responds more strongly to its own shaped direction than to a direction taken from another trajectory, even when both samples share a class label. Because this permutation preserves the population of shaped directions while breaking their pairing with the late state, the result shows that shared directional anisotropy alone cannot explain the effect.

We introduce a matched replay measurement for \emph{cross-time directional selection} in diffusion sampling. It isolates the relation between directions shaped by early updates and the sensitivity of later updates, beyond anisotropy measured at a single state. The matched and donor comparisons show both that early updates shape directions that are preferentially transmitted later and that this preference depends on their alignment with the receiving trajectory.

We also examine endpoint image-feature changes and report trajectory
geometry and fresh-probe crossover timing in Appendix~\ref{app:secondary}.

\begin{figure}[t]
  \centering
  \includegraphics[width=\linewidth]{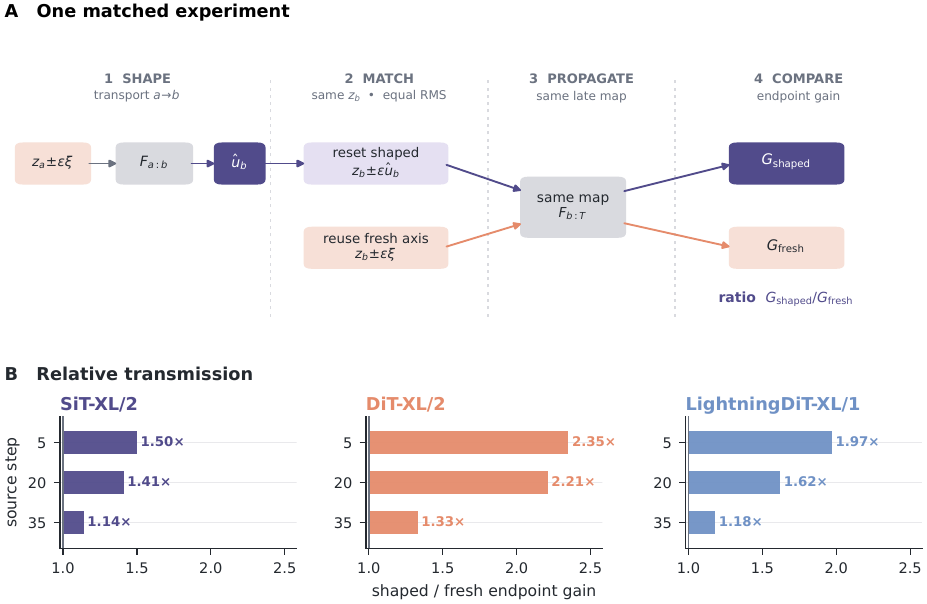}
  \caption{\textbf{Matched replay tests cross-time directional selection.}
  (A) We transport a perturbation from $a$ to $b$ to obtain an upstream-shaped
  direction. We then replay this direction and a fresh Gaussian direction at equal
  RMS magnitude around the same unperturbed late state $z_b$, and pass both through
  the same remaining updates.
  (B) Median paired shaped-to-fresh endpoint angular-gain ratios,
shown for each sampler and source time.}

  \label{fig:selective}
\end{figure}

\section{Related Work}
\label{sec:related}

\paragraph{Temporal structure and trajectory geometry.}
Diffusion sampling has distinct temporal regimes, including symmetry breaking,
speciation, and feature-decision windows
\citep{raya2023,biroli2024,sclocchi2024,li2024}. Sampling paths also exhibit regular geometric structure \citep{chen2024}, while derivatives of normalized score or update directions identify time ranges that are particularly sensitive to intervention \citep{sakamoto2026}.
These studies characterize how sampling changes over time; our alignment and angular-speed measurements situate the matched perturbation experiments within that temporal structure.

\paragraph{Error propagation and contraction.}
Local errors can accumulate, composed flow maps can amplify them, and stepwise
contraction guarantees can fail \citep{livanderschaar2024,gao2026,li2026}.
Spatial integration errors have also been tracked through the remaining
sampler \citep{yin2026}. These results motivate asking whether the downstream response also depends on the direction of the error. We hold the late state, perturbation
magnitude, and remaining updates fixed, then compare fresh isotropic directions with directions shaped by earlier sampling steps.

\paragraph{Directional geometry.}
Low-noise scores approximate normal directions to the data manifold
\citep{stanczuk2023}, and score-Jacobian spectra distinguish tangent and normal subspaces across generative phases \citep{ventura2025}. These results suggest that directional anisotropy may shape how perturbations propagate. Our experiment compares fresh and upstream-shaped directions under the same remaining updates. By transporting directions across time and exchanging them between trajectories, we test whether the resulting directional preference persists downstream and depends on the receiving state.

\section{Methods}
\label{sec:methods}

  \subsection{Setting and trajectory geometry}
  \label{sec:geometry-method}

  We analyze deterministic sampling trajectories from pretrained SiT-XL/2
  \citep{ma2024}, DiT-XL/2 \citep{peebles2023}, and LightningDiT-XL/1
  \citep{yao2025}, all at ImageNet $256\times256$ resolution with guidance scale
  1. SiT and LightningDiT use 49 uniform Euler updates, while DiT uses 50
  deterministic DDIM updates \citep{song2021}. We treat each checkpoint--sampler
  pair as a separate object of study. Let $z_i\in\mathbb{R}^d$ denote the
  flattened latent after update $i$, let $T$ denote the final update, and let
  $F_{i:j}$ denote the composed map from $i$ to $j$.

  We characterize each trajectory by its alignment with the final latent and its
  angular speed between adjacent updates, where $C$ denotes cosine similarity:
  \begin{equation}
    A_i=C(z_i,z_T),
    \qquad
    \omega_i=\frac{\arccos C(z_i,z_{i+1})}{\Delta\tau_i}.
    \label{eq:trajectory-geometry}
  \end{equation}

  Here $\Delta\tau_i=1/T$, and adjacent-update measurements are plotted at
  $(i+1/2)/T$. Thus $\omega_i$ measures rotation of the latent state about the
  origin, indexed by update progress. To separate this rotation from radial
  change, we project $\Delta z_i=z_{i+1}-z_i$ orthogonally to $z_i$ and compare
  $\omega_i$ with the relative tangential speed
  $q_i=\lVert\Delta z_i^{\mathrm{tan}}\rVert_2/
  (\lVert z_i\rVert_2\Delta\tau_i)$, which agrees with $\omega_i$ for small
  updates.

\subsection{Composed perturbation gain}
\label{sec:gain-method}

  The effect of a perturbation at a later state is determined by all subsequent updates. We therefore draw an isotropic Gaussian direction $\xi$ and normalize it so that $\RMS(\xi)=1$,
  where $\RMS(x)=\lVert x\rVert_2/\sqrt d$. At state $z_i$, we construct
  oppositely signed perturbations and propagate both through $F_{i:j}$, and define angular gain as the ratio of their final and
  initial angular separations:
  \begin{equation}
    z_i^{\pm}=z_i\pm r\RMS(z_i)\xi,
    \qquad
    z_j^{\pm}=F_{i:j}(z_i^{\pm}),
    \qquad
    G_{i:j}^{\theta}
    =\frac{\arccos C(z_j^+,z_j^-)}{\arccos C(z_i^+,z_i^-)}.
    \label{eq:angular-gain}
  \end{equation}

Here $r$ is the perturbation size relative to $\RMS(z_i)$. $G_{i:j}^{\theta}>1$ indicates that the remaining map increases angular
  separation, while $G_{i:j}^{\theta}<1$ indicates contraction. Using opposite
  signs gives a symmetric finite-radius measurement and reduces one-sided
  nonlinear effects. We also measure latent-RMS gain:
  \begin{equation}
    G_{i:j}^{\mathrm{RMS}}=
    \frac{\RMS(z_j^+-z_j^-)}{\RMS(z_i^+-z_i^-)}.
    \label{eq:rms-gain}
  \end{equation}
  Angular gain measures separation in latent direction, whereas RMS gain measures
  Euclidean separation in latent space. Their relation is given below.

\subsection{Matched fresh and upstream-shaped directions}
\label{sec:matched-method}

At a fixed late boundary $b$, we compare the fresh probe $\xi$ with the same
probe introduced at an earlier state $z_a$ and shaped by transport to $b$:
\begin{align}
  d_b^{(a)}
  &=\tfrac12\!\left[
    F_{a:b}\!\left(z_a+r_a\RMS(z_a)\xi\right)
    -F_{a:b}\!\left(z_a-r_a\RMS(z_a)\xi\right)
  \right],
  &u_b^{(a)}&=\frac{d_b^{(a)}}{\RMS(d_b^{(a)})}, \notag\\[-2pt]
  \widetilde z_b^{k,\pm}
  &=z_b\pm r_b\RMS(z_b)u_b^k,
  &u_b^{\mathrm{fresh}}&=\xi.
  \label{eq:matched-construction}
\end{align}

We discard the transported branches before replay and return to the unperturbed
  state $z_b$. With $k\in\{\mathrm{fresh},\mathrm{shaped}(a)\}$, both probes
  therefore begin at the same state, have the same RMS radius, and undergo the
  same map $F_{b:c}$ (Figure~\ref{fig:selective}A). We set
  $r_a=r_b=0.00125$, use $a\in\{5,20,35\}$ and $b=40$, and evaluate $c=44$
  and $c=T$. Reusing the fresh axis across source conditions gives each shaped
  direction the same paired fresh baseline.

\paragraph{Local interpretation and choice of metric.}
  Let $J_{i:j}$ denote the derivative of the composed numerical map, evaluated
  at the unperturbed trajectory. As the source radius tends to zero, the shaped
  direction is proportional to $J_{a:b}\xi$. The corresponding downstream RMS
  gains compare
  $\lVert J_{b:c}u_b^{(a)}\rVert/\lVert u_b^{(a)}\rVert$ and
  $\lVert J_{b:c}\xi\rVert/\lVert\xi\rVert$.

  Angular gain instead depends on the components tangent to the latent sphere.
  Writing $\widehat z_i=z_i/\lVert z_i\rVert$ and
  $P_i=I-\widehat z_i\widehat z_i^\top$, its local expression is
  \begin{equation}
   G_{b:c}^{\theta}(u)\simeq
   \frac{\lVert z_b\rVert}{\lVert z_c\rVert}
   \frac{\lVert P_cJ_{b:c}u\rVert}{\lVert P_bu\rVert}.
   \label{eq:angular-local}
  \end{equation}
  Equal RMS perturbations can have different initial angular separations.
  Accordingly, an angular gain below one establishes angular contraction only.
  The paired RMS comparison determines whether the directional ordering also
  appears in latent-space distance.

\paragraph{Sampling and aggregation.}
For each model, we sample 1,024 independently seeded trajectories with labels cycling through the 1,000 ImageNet classes (24 classes occur twice and the remainder once). We evaluate two Gaussian axes per trajectory, with the positive and negative branches forming one symmetric measurement. For trajectory $n$, we take the median across axes separately for
shaped and fresh gains, yielding $g_n^{(a)}$ and $g_n^f$, and compute the paired ratio $g_n^{(a)}/g_n^f$. We report the median of these paired ratios across trajectories, separately for each model. This differs from the ratio of marginal median gains.

\paragraph{Donor replay.}
We run donor replay on new 256-trajectory cohorts for SiT and DiT, with four seeds for each of 64 ImageNet classes. For every source condition, we construct two within-class and two across-class permutations of shaped directions and replay each donor direction around its recipient's late state. The permutation preserves the full shaped-direction ensemble while changing its pairing with the recipient state. Class conditioning, replay magnitude, late boundary, and remaining updates are held fixed. We summarize donor-to-fresh and own-to-donor paired gain ratios using the aggregation above.

\subsection{Endpoint representation sensitivity}
\label{sec:endpoint-method}

The gain experiment measures how perturbations propagate through the late sampler. A second experiment asks whether the resulting directions differ in their decoded effect once their magnitude is equalized. For each fresh or shaped direction at $T$, we normalize the centered branch displacement to unit
RMS and replay it at radius $0.00125\RMS(z_T)$ around the unperturbed endpoint. No diffusion updates follow. We decode the plus and minus replay states with the same VAE and use half their
pixel-RMS distance and half their feature-space distance as responses. We use L2-normalized DINOv2 CLS tokens and mean-pooled patch tokens \citep{oquab2023}, as well as ResNet-50 features \citep{he2016}.

For each response $H$, we compute the shaped-to-fresh ratio for each probe axis, take the geometric mean of the two axis-wise ratios within each trajectory, and report the median of these trajectory-level ratios. This aggregation differs from the gain comparison above, which first
takes the median gain across axes separately for shaped and fresh probes
and then forms their ratio within each trajectory. We also report the fraction of axis pairs with a larger shaped response;
these pairs share trajectories and are descriptive rather than independent observations. Together, the pixel and feature responses measure the effect of equal-magnitude endpoint perturbations on decoded images and image
representations. We run this endpoint experiment on 256 SiT trajectories.
  
\section{Results}
\label{sec:results}



Our main result compares upstream-shaped and fresh directions at the same late state. We report the size of this effect, its robustness checks, and the donor control. Finally, endpoint replay tests whether equal-magnitude directions produce different changes in decoded image features. Native trajectory geometry and fresh-probe crossover timing are reported in Appendix C.

\subsection{Late updates preferentially transmit upstream-shaped directions}
\label{sec:matched-results}

At the late boundary, fresh and upstream-shaped perturbations are injected at the same RMS magnitude around the same unperturbed state and undergo the same remaining updates. Across every model and source time, the upstream-shaped direction produces greater endpoint angular gain (Figure~\ref{fig:selective}; Table~\ref{tab:selective}).

\begin{table}[t]
  \centering
\caption{\textbf{Median endpoint angular gain.} Parentheses give the median
  trajectory-level shaped-to-fresh gain ratio. Each entry summarizes 1,024
  trajectories with two probe axes per trajectory.}
  \label{tab:selective}
  \small
  \begin{tabular}{lcccc}
    \toprule
    Model & Fresh & Shaped from 5 & Shaped from 20 & Shaped from 35 \\
    \midrule
    SiT-XL/2
      & 0.816 & 1.202 (1.496$\times$) & 1.132 (1.409$\times$) & 0.927 (1.138$\times$) \\
    DiT-XL/2
      & 0.766 & 1.725 (2.350$\times$) & 1.638 (2.214$\times$) & 1.017 (1.330$\times$) \\
    LightningDiT-XL/1
      & 0.560 & 1.096 (1.969$\times$) & 0.908 (1.623$\times$) & 0.660 (1.179$\times$) \\
    \bottomrule
  \end{tabular}
\end{table}

For every model and source time, shaped gain exceeds its matched fresh gain for all 1,024 trajectories. The ordering is already present over the four-update window from the replay boundary, where median shaped-to-fresh ratios range from 1.079 to 1.552.

The effect is largest for early source times. Fresh probes have median angular gain below one in all three models, whereas directions shaped from step 5 have median gain above one. The advantage declines as the source approaches the
replay boundary. At step 35, shaped probes also contract in SiT and LightningDiT, but less than their matched fresh probes. Thus the late map
responds more strongly to upstream-shaped directions whether it expands or
contracts the perturbation.

The paired contrast is stable across the tested replay radii. In the SiT audit, an eight-fold change in replay radius alters the median paired ratio by at most 0.0004 (supplementary Figure~\ref{fig:radius-audit}). A separate source-radius audit changes the median paired angular-gain ratio by less than $0.009\%$ in every model (Appendix~\ref{app:ready-followups}).

Latent-RMS gain preserves the shaped-over-fresh ordering, although whether an individual gain lies above or below one can depend on the metric. For example, SiT directions shaped from step 35 have median angular gain 0.927 and median RMS gain 1.083; for LightningDiT directions shaped from step 20, the corresponding values are 0.908 and 1.084. The robust result is stronger relative transmission of shaped directions. In these cases, angular separation contracts while latent-space distance expands.

\subsection{The advantage depends on pairing directions with their trajectory}
\label{sec:donors}

We evaluate donor replay on new 256-trajectory cohorts for SiT and DiT. Shaped
  directions are reassigned to recipient trajectories either within class or
  across classes, while the recipient state, conditioning, replay magnitude, and
  remaining updates remain fixed.

  At source step 5, own shaped directions have median gain ratios of 1.509 in
  SiT and 2.353 in DiT relative to fresh probes. Reassigned donor directions
  retain a substantially smaller advantage: donor-to-fresh ratios are 1.059 and 1.191 for
  within-class assignments, and 1.057 and 1.192 for across-class assignments.
  Direct own-to-donor ratios are 1.413 and 2.004 within class, and 1.410 and
  2.001 across classes. The own-direction advantage remains at sources 20 and
  35 (Appendix~\ref{app:ready-followups}).

  Reassigning directions preserves their population while substantially reducing
  their advantage. The result therefore depends on how a shaped direction aligns
  with its recipient trajectory, beyond shared or class-level directional
  structure.

This dependence on direction–trajectory pairing is also compatible with trajectory-specific correlations between upstream and downstream Jacobians in the small-perturbation regime.

\subsection{Early-shaped directions induce larger endpoint feature responses}
\label{sec:endpoint-results}

The matched-gain result shows that upstream-shaped directions propagate more
  strongly through the late sampler. We next ask whether equal-magnitude replay
  of these directions also produces different decoded responses. In SiT,
  directions shaped from step 5 produce larger DINOv2 and ResNet-50 feature
  responses than their paired fresh directions, while the pixel-RMS response
  ratio remains near one (Table~\ref{tab:endpoint}; supplementary
  Figure~\ref{fig:endpoint-secondary}).

\begin{table}[t]
  \centering
    \caption{\textbf{Shaped-to-fresh response ratios after equal-RMS endpoint replay in SiT.} Values are medians across 256 trajectories after geometrically combining the two axis-wise ratios within each trajectory. Values above one indicate a larger response to the shaped direction.}

  \label{tab:endpoint}
  \small
  \begin{tabular}{lccc}
    \toprule
    Response & Shaped from 5 & Shaped from 20 & Shaped from 35 \\
    \midrule
    Pixel RMS          & 1.007 & 0.959 & 0.964 \\
    DINOv2 CLS         & 1.264 & 1.116 & 0.999 \\
    DINOv2 mean patch  & 1.256 & 1.156 & 1.017 \\
    ResNet-50          & 1.135 & 1.066 & 0.990 \\
    \bottomrule
  \end{tabular}
\end{table}

At source step 5, shaped responses exceed fresh responses for 76.6\% of axis
  pairs in DINOv2 CLS space, 79.5\% in mean DINOv2 patch space, and 67.8\% in
  ResNet-50 space, compared with 50.0\% in pixel RMS. The feature-space
  advantage weakens at source step 20 and is absent or nearly absent at step 35.
  Because every direction is replayed at the same RMS magnitude around the same
  endpoint state, inherited transport magnitude cannot explain this contrast.
  The effect therefore extends from latent-space propagation to decoded
  image-feature representations.

\section{Discussion and Limitations}
\label{sec:discussion}
The matched experiment separates the response to newly drawn isotropic directions from the response to directions produced by earlier updates. With the late state and replay magnitude held fixed, these direction populations receive different downstream gain. Fresh-probe averages therefore describe the response to a particular probe
population when assessing deviations carried through sampling. The Gaussian
probes serve as controlled interventions. Their relation to solver,
quantization, and caching error distributions is not evaluated in the
reported experiments.



\paragraph{Interpretation of the contrast.}
In the small-perturbation regime, upstream transport approximately
transforms a probe by $J_{a:b}$, while the remaining updates act through
$J_{b:c}$. Correlations between the sensitive directions of these
trajectory-dependent maps could therefore account for the shaped-direction
advantage. The stability of the paired ratio over the tested small-radius
range is consistent with this interpretation. Donor reassignment shows
that pairing with the recipient trajectory matters, but remains compatible
with the same linear account. The reported experiments characterize this
directional preference without isolating its underlying mechanism.


\paragraph{Measurement limits.}
The paired summaries use two probe axes per trajectory and characterize typical directional response rather than local spectra or worst-case gain. They concern discrete samplers at fixed step counts. Solver choice, stochastic sampling, guidance scale, and broader model families define the current scope. Normalized update index aligns plotting positions across models, while their noise schedules, latent representations, and training recipes differ.

\paragraph{Endpoint and timing interpretations.}
The endpoint experiment is a SiT study with a fixed VAE and feature encoders. It measures representation response under equal-magnitude replay rather than human-perceptual fidelity. The timing association describes co-variation across trajectories; it does not identify the cause of the matched contrast.

These results matter for evaluating sampler approximations. A late window can attenuate fresh perturbations while continuing to transmit directions arriving from earlier updates. Evaluations of reduced precision or cached computation therefore need perturbations that reflect this upstream transport. Matched
replay controls their arrival magnitude, and pairing it with measured approximation residuals would test consequences for endpoint fidelity and
sample quality.

\section{Conclusion}
\label{sec:conclusion}

Across the three samplers, directions shaped by earlier updates receive greater late-map gain than fresh isotropic directions replayed at the same state and RMS magnitude. The shaped-over-fresh ordering appears in both angular and
latent-RMS measurements, while the location of unit gain depends on the metric. In SiT and DiT, reassigning shaped directions to other trajectories substantially reduces their advantage, including within class, showing that
direction--trajectory pairing matters.

A stability measurement is defined jointly by its time window and perturbation population. Tracking directions through the preceding sampler reveals a late response that fresh probes alone do not capture, and shows that much of the advantage is tied to the trajectory on which those directions were formed.

\section{Impact Statement}
\label{sec:impact}

This work studies the sampling dynamics of existing image generators and does not introduce a new generation or editing capability. The diagnostic itself requires additional perturbed sampling branches. If used to guide approximate
sampling or training changes, it could reduce generation cost while inheriting the risks of the underlying image generator. Such use should be evaluated for image quality, diversity, and bias.

\section*{Acknowledgements}
\label{sec:ack}
We thank CloudRift for providing the computational resources that supported
the experiments in this work.

  \appendix
  \section{Experimental details}
  \label{app:details}

  All models generate class-conditional ImageNet samples at $256\times256$
  resolution with guidance scale 1. SiT-XL/2 and LightningDiT-XL/1 use 49
  fixed-grid Euler updates, while DiT-XL/2 uses 50 DDIM updates. The main replay
  boundary is $b=40$, source steps are $a\in\{5,20,35\}$, and the short window
  ends at step 44. The main source and replay radii are both 0.00125 relative to
  the RMS magnitude of the corresponding latent state.

  The main comparison uses 1,024 trajectories and two independent Gaussian axes
  per trajectory. Positive and negative branches form one centered probe. We run
  sampling and perturbation propagation in FP32 with TF32 disabled. For small
  angular separations, we compute the unit-sphere chord $c$ and recover
  $\theta=2\arcsin(c/2)$ for numerical stability.

  Reported 95\% percentile intervals use 10,000 paired trajectory bootstrap
  draws, retaining all axes and source conditions from a trajectory in each draw.
  Donor intervals resample classes, with four trajectories per class and fixed
  donor assignments; these intervals are descriptive because donor reuse induces
  dependence between recipients. We do not pool results across models.

  For the replay-radius audit, we evaluate
  $\{0.0003125,0.000625,0.00125,0.0025\}$ on 256 SiT trajectories while fixing
  the source radius at 0.00125. The source-radius audit varies the source radius
  over $\{0.000625,0.00125,0.0025\}$ while holding the replay radius at 0.00125.

  The endpoint experiment uses 256 SiT trajectories, the
  \path{stabilityai/sd-vae-ft-mse} decoder, \path{facebook/dinov2-base}, and
  ImageNet-1K V2 ResNet-50 weights. Pixel RMS is computed on raw decoded images.
  Encoder inputs are mapped to $[0,1]$, clamped, resized, center-cropped, and
  normalized. DINOv2 uses bicubic resize and ResNet-50 uses bilinear resize, both
  with antialiasing.

  \begin{figure}[t]
    \centering
    \includegraphics[width=0.58\linewidth]{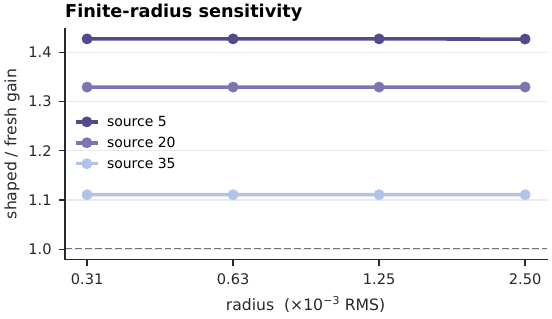}
    \caption{\textbf{Finite-radius control in SiT.}
    On a separate 256-trajectory cohort, the median paired shaped-to-fresh
    endpoint angular-gain ratio changes by at most 0.0004 across the
    eightfold replay-radius range, with the source radius fixed.}
    \label{fig:radius-audit}
  \end{figure}

  \begin{table}[ht]
  \centering
  \caption{\textbf{Median endpoint latent-RMS gain under matched replay.}
  The shaped-over-fresh ordering agrees with Table~\ref{tab:selective}, although
  angular and latent-RMS gain can fall on opposite sides of one.}
  \small
  \begin{tabular}{lrrrr}
  \toprule
  Model & Fresh & Source 5 & Source 20 & Source 35 \\
  \midrule
  SiT-XL/2 & 0.952 & 1.403 & 1.324 & 1.083 \\
  DiT-XL/2 & 0.723 & 1.627 & 1.540 & 0.953 \\
  LightningDiT-XL/1 & 0.668 & 1.307 & 1.084 & 0.786 \\
  \bottomrule
  \end{tabular}
  \end{table}

  \section{Focused follow-up measurements}
  \label{app:ready-followups}

  \paragraph{Donor replay.}
  Donor replay uses new 256-trajectory cohorts for SiT and DiT, with four seeds
  for each of 64 ImageNet classes. For every source step, two within-class and
  two across-class permutations reassign shaped directions to recipient states.
  The permutations preserve the shaped-direction population while changing its
  pairing with the recipient. Recipient conditioning, replay magnitude, late
  boundary, and remaining updates remain fixed. Intervals use 10,000
  class-block bootstrap draws with donor assignments held fixed. Because donors
  are reused across recipients, these intervals are descriptive.

  \begin{table}[ht]
  \centering
  \small
  \caption{\textbf{Median paired endpoint angular-gain ratios for donor replay.}
  Parentheses give 95\% class-block bootstrap intervals.}
  \begin{tabular}{llccc}
  \toprule
  Model & Source & Own/fresh & Own/within & Own/across \\
  \midrule
  SiT & 5  & 1.509 (1.468--1.531) & 1.413 (1.388--1.447) & 1.410 (1.381--1.447) \\
  SiT & 20 & 1.413 (1.390--1.444) & 1.357 (1.329--1.381) & 1.354 (1.329--1.383) \\
  SiT & 35 & 1.141 (1.134--1.151) & 1.121 (1.111--1.131) & 1.121 (1.114--1.130) \\
  DiT & 5  & 2.353 (2.279--2.504) & 2.004 (1.906--2.094) & 2.001 (1.913--2.105) \\
  DiT & 20 & 2.255 (2.192--2.379) & 2.050 (1.922--2.141) & 2.017 (1.942--2.114) \\
  DiT & 35 & 1.337 (1.323--1.364) & 1.284 (1.270--1.313) & 1.293 (1.273--1.325) \\
  \bottomrule
  \end{tabular}
  \end{table}

  \paragraph{Source-radius audit.}
  The audit uses 128 trajectories per model across 32 classes, including
  LightningDiT. It varies the source radius over
  $\{0.000625,0.0025\}$ around the baseline 0.00125 while keeping the replay
  radius fixed at 0.00125. The table reports the largest absolute change in the
  median paired angular-gain ratio across source steps 5, 20, and 35.

  \begin{table}[ht]
  \centering
  \small
  \caption{\textbf{Stability of the paired angular-gain ratio under source-radius
  variation.}}
  \begin{tabular}{lr}
  \toprule
  Model & Largest absolute change in median paired ratio \\
  \midrule
  SiT & 0.007\% \\
  DiT & 0.008\% \\
  LightningDiT & 0.004\% \\
  \bottomrule
  \end{tabular}
  \end{table}

  The fifth-percentile direction cosines across source-radius comparisons are
  0.9974 for SiT, 0.9978 for DiT, and 0.9932 for LightningDiT.

\paragraph{Numerical accuracy.}
To assess numerical error at the tested perturbation scales, we compare
zero-perturbation continuations across microbatch sizes. The largest
discrepancy, normalized by the smallest tested probe endpoint response,
is 0.385\% for SiT, 0.897\% for DiT, and 1.133\% for LightningDiT.
Zero-perturbation endpoint discrepancies are below
$3.67\times10^{-6}$ in relative RMS. These discrepancies are small
compared with the measured perturbation responses.


\subsection{Uncertainty for the main cohorts}

  \begin{table}[ht]
  \centering
  \small
  \caption{\textbf{Paired trajectory bootstrap intervals for the main
  comparison.} Entries are 95\% intervals for the median shaped-to-fresh endpoint
  angular-gain ratios in Table~\ref{tab:selective}, based on 10,000 draws.}
  \begin{tabular}{lccc}
  \toprule
  Model & Source 5 & Source 20 & Source 35 \\
  \midrule
  SiT & [1.482, 1.515] & [1.399, 1.421] & [1.133, 1.142] \\
  DiT & [2.290, 2.409] & [2.153, 2.259] & [1.322, 1.342] \\
  LightningDiT & [1.951, 1.983] & [1.609, 1.635] & [1.176, 1.181] \\
  \bottomrule
  \end{tabular}
  \end{table}

  For source-5 endpoint replay, the corresponding 95\% ratio intervals are
  [0.986, 1.019] for pixel RMS, [1.180, 1.337] for DINOv2 CLS,
  [1.224, 1.313] for mean DINOv2 patch features, and [1.104, 1.162] for
  ResNet-50. The pixel-RMS interval includes one.

  For the timing cohort, the 95\% interval for median crossover progress is
  [0.429, 0.461]. The interval for the median difference between the
  trajectory-level angular-speed peak and crossover is [0.178, 0.200]. The
  Spearman correlation between these quantities is $\rho=0.622$, with interval
  [0.536, 0.698]. These intervals quantify sampling variation for the fixed
  models, boundaries, and estimands studied here.

\section{Secondary analyses}
\label{app:secondary}

\begin{figure}[t]
  \centering
  \includegraphics[width=0.9\linewidth]{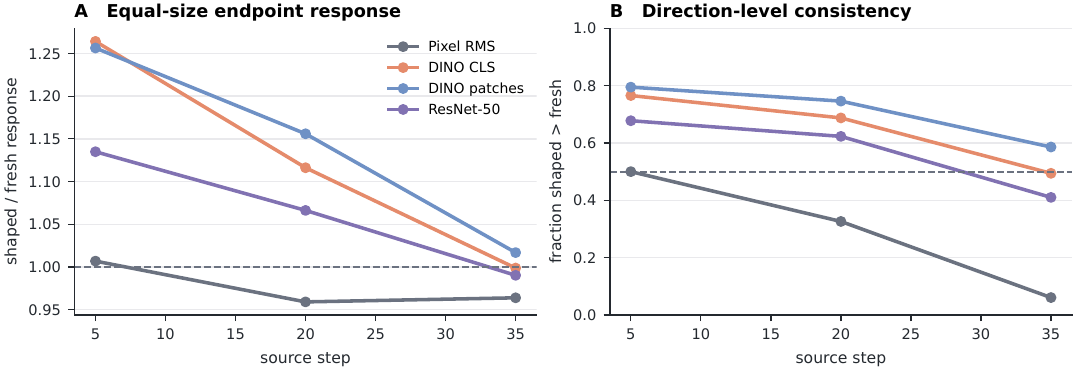}
  \caption{\textbf{Equal-RMS endpoint replay produces different feature
  responses.} (A) Median shaped-to-fresh response ratios after replaying both
  directions at equal RMS magnitude around the same endpoint state. (B) Fraction
  of matched axis pairs for which the shaped response exceeds the fresh response.}
  \label{fig:endpoint-secondary}
\end{figure}

\subsection{Timing of angular motion and fresh-probe contraction}

We examine how the full-horizon angular gain of fresh probes changes as
their injection time moves later in sampling, and compare the transition
to contraction with the timing of angular motion along the unperturbed
trajectory.

We evaluate fresh probes injected at steps
$\{4,8,12,16,20,24,28,32,36,40,44\}$ on 256 SiT trajectories. For each
trajectory, the crossover is the first injection whose median gain across axes
falls below one and remains below one at all later tested injections. We
estimate its location by linearly interpolating log gain around the crossing
and normalize it by 49; persistence is assessed on the tested injection grid.

In this cohort, median crossover progress is 0.451 and median angular-speed
peak progress is 0.643. The two quantities have Spearman correlation
$\rho=0.622$ across trajectories; the axis-specific correlations are 0.626
and 0.555. Trajectories with later angular-speed peaks also tend to have later
fresh-probe crossovers, while contraction typically begins before peak angular
motion. Thus, angular contraction of nearby perturbations can coexist with
substantial angular motion of the underlying trajectory. This timing
association is descriptive and does not establish a mechanism for the
shaped-to-fresh gain contrast.

These timing measurements use a separate 256-trajectory SiT cohort from
the 1,024-trajectory cohorts used to characterize geometry below.

\begin{figure}[t]
  \centering
  \includegraphics[width=0.72\linewidth]{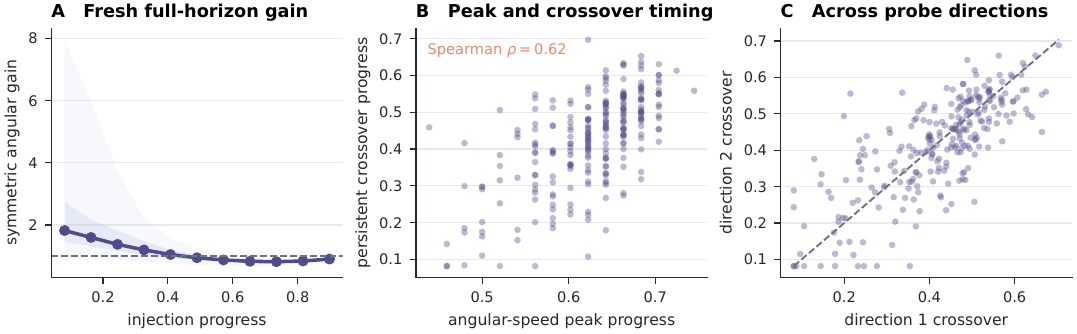}
  \caption{\textbf{Timing association in SiT.} Fresh-probe crossover time and
  angular-speed peak time co-vary across 256 trajectories, with contraction
  typically preceding peak angular motion.}
  \label{fig:timing-secondary}
\end{figure}

\subsection{Trajectory geometry and tangential-speed validation}
\label{sec:geometry-results}

We characterize unperturbed trajectory geometry on 1,024 trajectories per
model using the measurements defined in Section~3.1.

\begin{figure}[t]
  \centering
  \includegraphics[width=\linewidth]{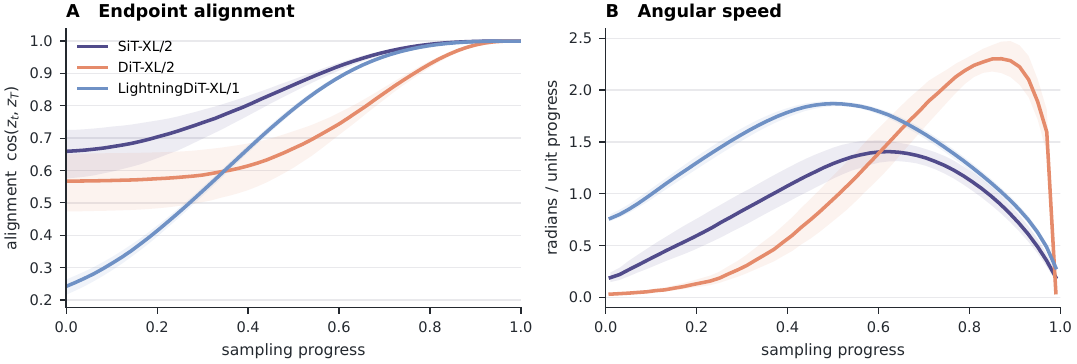}
  \caption{\textbf{Alignment and angular motion capture different aspects of
  sampling.} (A) Median alignment between each latent state and its endpoint
  increases smoothly. (B) Median angular speed of the latent state direction
  is non-monotonic and peaks at a model-dependent time. Shading denotes the
  interquartile range across 1,024 trajectories per model.}
  \label{fig:geometry}
\end{figure}

Alignment with the endpoint rises smoothly in all three models
(Figure~\ref{fig:geometry}A). Meanwhile, angular speed rises to a
model-specific maximum before falling (Figure~\ref{fig:geometry}B). The median
position of the trajectory-level angular-speed peak is normalized progress
0.622 for SiT, 0.870 for DiT, and 0.500 for LightningDiT. Thus, a state can
become increasingly aligned with its endpoint while continuing to rotate
substantially.

The relative tangential-speed control closely tracks the angular-speed
profile. Their median curves differ by at most 0.019 for SiT, 0.018 for DiT,
and 0.024 for LightningDiT, and their median trajectory-level peak-step
difference is zero in every model. Absolute tangential motion also peaks near
the angular maximum. These controls support that the angular-speed peak
reflects increased sideways motion, rather than merely normalization by a
changing latent norm.

\subsection{Token norm has little association with the gain contrast}

On 256 SiT trajectories, we correlate token norm immediately before
perturbation with token-level endpoint log gain. The median within-trajectory
Spearman correlation has magnitude below 0.06 for every condition, and every
interquartile range contains zero. Correlations between boundary token norm and
the shaped-minus-fresh gain advantage are $-0.081$, $-0.058$, and $-0.079$ for
source steps 5, 20, and 35. Population-median gain also varies little across
token locations. Token norm therefore explains little of the measured gain
contrast.

\begin{figure}[t]
  \centering
  \includegraphics[width=0.92\linewidth]{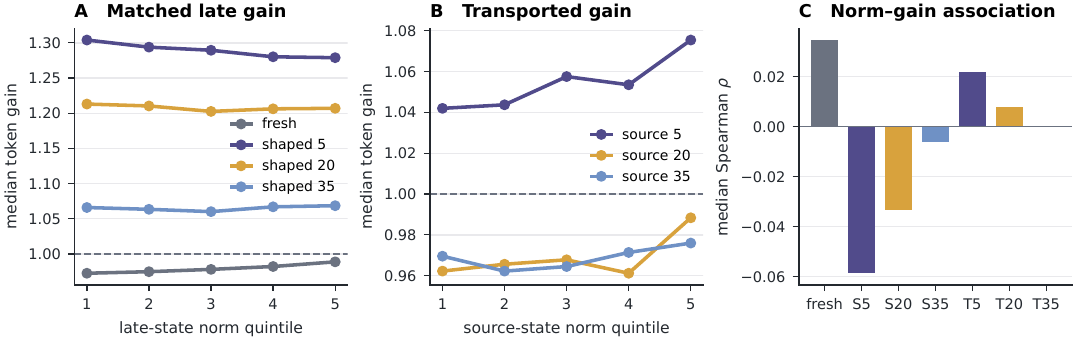}
  \caption{\textbf{Token norm and local gain in SiT.} Pre-perturbation token
  norm has little consistent association with fresh, reset-shaped, or
  transported endpoint gain.}
  \label{fig:token-secondary}
\end{figure}

\begin{thebibliography}{99}
\small

\bibitem[Biroli et~al.(2024)Biroli, Bonnaire, de~Bortoli, and M\'ezard]{biroli2024}
G.~Biroli, T.~Bonnaire, V.~de~Bortoli, and M.~M\'ezard. Dynamical regimes of
diffusion models. \emph{Nature Communications}, 15:9957, 2024.

\bibitem[Chen et~al.(2024)Chen, Zhou, Wang, Shen, and Lyu]{chen2024}
D.~Chen, Z.~Zhou, C.~Wang, C.~Shen, and S.~Lyu. On the trajectory regularity of
ODE-based diffusion sampling. \emph{ICML}, 2024.

\bibitem[Gao et~al.(2026)Gao, Li, Shi, and Zhou]{gao2026}
W.~Gao, M.~Li, L.~Shi, and H.~Zhou. A quantitative approximation framework for
flow distillation in diffusion models. \emph{arXiv:2606.03820}, 2026.

\bibitem[Li and Chen(2024)]{li2024}
M.~Li and S.~Chen. Critical windows: Non-asymptotic theory for feature emergence
in diffusion models. \emph{ICML}, 2024.

\bibitem[Li and van~der Schaar(2024)]{livanderschaar2024}
Y.~Li and M.~van~der Schaar. On error propagation of diffusion models.
\emph{ICLR}, 2024.

\bibitem[Li and Zeng(2026)]{li2026}
S.~Li and D.~Zeng. Mitigating the contractivity trap in diffusion ODEs via
Stein stabilization. \emph{ICML}, 2026.

\bibitem[Raya and Ambrogioni(2023)]{raya2023}
G.~Raya and L.~Ambrogioni. Spontaneous symmetry breaking in generative diffusion
models. \emph{NeurIPS}, 2023.

\bibitem[Sakamoto and Sakamoto(2026)]{sakamoto2026}
R.~Sakamoto and K.~Sakamoto. The geometry of phase transitions in generative
dynamics via projection caustics. \emph{arXiv:2606.13191}, 2026.

\bibitem[Sclocchi et~al.(2024)Sclocchi, Favero, and Wyart]{sclocchi2024}
A.~Sclocchi, A.~Favero, and M.~Wyart. A phase transition in diffusion models
reveals the hierarchical nature of data. \emph{PNAS}, 121:e2408799121, 2024.

\bibitem[Stanczuk et~al.(2023)Stanczuk, Batzolis, Deveney, and Sch\"onlieb]{stanczuk2023}
J.~Stanczuk, G.~Batzolis, T.~Deveney, and C.-B.~Sch\"onlieb. Your diffusion
model secretly knows the dimension of the data manifold.
\emph{arXiv:2212.12611}, 2023.

\bibitem[Ventura et~al.(2025)Ventura, Achilli, Silvestri, Lucibello, and Ambrogioni]{ventura2025}
E.~Ventura, B.~Achilli, G.~Silvestri, C.~Lucibello, and L.~Ambrogioni.
Manifolds, random matrices and spectral gaps: The geometric phases of generative
diffusion. \emph{arXiv:2410.05898}, 2025.

\bibitem[Yin(2026)]{yin2026}
S.~Yin. Spatial transport of integration error in generative ODEs.
\emph{arXiv:2607.16361}, 2026.

\bibitem[Ma et~al.(2024)]{ma2024}
N.~Ma, M.~Goldstein, M.~S.~Albergo, N.~M.~Boffi, E.~Vanden-Eijnden, and
S.~Xie. SiT: Exploring flow and diffusion-based generative models with scalable
interpolant transformers. \emph{ECCV}, 2024.
\url{https://arxiv.org/abs/2401.08740}.

\bibitem[Peebles and Xie(2023)]{peebles2023}
W.~Peebles and S.~Xie. Scalable diffusion models with transformers.
\emph{ICCV}, 2023. \url{https://arxiv.org/abs/2212.09748}.

\bibitem[Yao et~al.(2025)]{yao2025}
J.~Yao, B.~Yang, and X.~Wang. Reconstruction vs. generation: Taming
optimization dilemma in latent diffusion models. \emph{CVPR}, 2025.
\url{https://arxiv.org/abs/2501.01423}.

\bibitem[Song et~al.(2021)]{song2021}
J.~Song, C.~Meng, and S.~Ermon. Denoising diffusion implicit models.
\emph{ICLR}, 2021. \url{https://arxiv.org/abs/2010.02502}.

\bibitem[Oquab et~al.(2023)]{oquab2023}
M.~Oquab et~al. DINOv2: Learning robust visual features without supervision.
\emph{arXiv:2304.07193}, 2023. \url{https://arxiv.org/abs/2304.07193}.

\bibitem[He et~al.(2016)]{he2016}
K.~He, X.~Zhang, S.~Ren, and J.~Sun. Deep residual learning for image recognition.
\emph{CVPR}, 2016. \url{https://arxiv.org/abs/1512.03385}.

\end{thebibliography}
\end{document}